\pdfoutput=1
\documentclass[11pt,table,xcdraw]{article}

\usepackage[]{acl}

\usepackage{times}
\usepackage{latexsym}
\usepackage{amsmath}

\usepackage[T1]{fontenc}
\usepackage[utf8]{inputenc}

\usepackage{microtype}

\usepackage{inconsolata}

\usepackage{graphicx}
\usepackage{subcaption}
\usepackage{amsfonts}
\usepackage{booktabs}
\usepackage{multirow}
\usepackage{colortbl}
\usepackage{adjustbox}

\title{OSoRA: Output-Dimension and Singular-Value Initialized Low-Rank Adaptation}

\author{
 \textbf{Jialong Han\textsuperscript{1,2}},
 \textbf{Si Zhang\textsuperscript{2}},
 \textbf{Ke Zhang\textsuperscript{2}},
\\
\\
 \textsuperscript{1}SIST, ShanghaiTech University\\
 \textsuperscript{2}SKLP, Institute of Computing Technology, Chinese Academy of Sciences
}

\begin{document}
\maketitle
\begin{abstract}
  Fine-tuning Large Language Models (LLMs) has become increasingly challenging due to their massive scale and associated computational costs.
  Parameter-Efficient Fine-Tuning (PEFT) methodologies have been proposed as computational alternatives; 
  however, their implementations still require significant resources.
  In this paper, we present OSoRA (Output-Dimension and Singular-Value Initialized Low-Rank Adaptation), 
  a novel PEFT method for LLMs.
  OSoRA extends Low-Rank Adaptation (LoRA) by integrating Singular Value Decomposition (SVD) 
  with learnable scaling vectors in a unified framework.
  It first performs an SVD of pre-trained weight matrices, then optimizes an output-dimension vector during training, 
  while keeping the corresponding singular vector matrices frozen.
  OSoRA substantially reduces computational resource requirements by minimizing the number of trainable parameters during fine-tuning.
  Comprehensive evaluations across mathematical reasoning, common sense reasoning, 
  and other benchmarks demonstrate that OSoRA achieves comparable or superior performance to state-of-the-art methods like LoRA and VeRA,
  while maintaining a linear parameter scaling even as the rank increases to higher dimensions.
  Our ablation studies further confirm that jointly training both the singular values and the output-dimension vector is critical for optimal performance.
\end{abstract}

\begin{figure}
  \centering
  \adjustbox{max width=\linewidth, clip}{
    \includegraphics{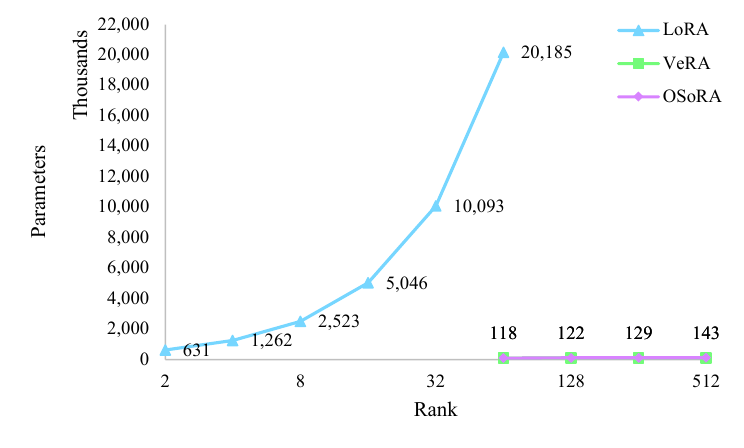}
  }
  \caption{Parameter count comparison among adaptation methods at varying ranks on Qwen2-7B model. 
  The results demonstrate that LoRA exhibits exponential growth in trainable parameters with increasing rank, 
  whereas both VeRA and OSoRA maintain efficient linear scaling in their parameter count.}
  \label{fig:gpqa}
\end{figure}

\section{Introduction}
Large Language Models (LLMs) have demonstrated remarkable capabilities across various Natural Language Processing (NLP) tasks. 
However, as these models escalate in size to hundreds of billions of parameters, 
fine-tuning them requires prohibitive computational resources \citep{medec2025,language2020}. 
This computational challenge has catalyzed the development of Parameter-Efficient Fine-Tuning (PEFT) methodologies,
which enable fine-tuning LLMs by selectively updating only a minimal subset of parameters.

Recent PEFT approaches include Low-Rank Adaptation (LoRA) \citep{lora2022}, which constrains weight updates to low-rank decompositions; 
Vector-based Random Matrix Adaptation (VeRA) \citep{vera2024}, which further improves efficiency by training only scaling vectors; 
and PiSSA \citep{pissa2024}, which uses Singular Value Decomposition (SVD) to update important weight matrix components. 
Despite these advances, existing methods still face limitations in balancing parameter efficiency with adaptation quality.

We introduce OSoRA (Output-Dimension and Singular-Value Initialized Low-Rank Adaptation), 
a novel PEFT method that combines SVD-based decomposition with a learnable scaling vector. 
OSoRA decomposes pretrained weight matrices using SVD, 
then selectively updates only the singular values and a single output-dimension vector during training.
This approach significantly reduces trainable parameters while maintaining competitive performance.

Our contributions include:
\begin{itemize}
  \item A novel PEFT method combining SVD-based decomposition with a learnable scaling vector
  \item Demonstration that updating only singular values and a single vector is sufficient for effective adaptation
  \item Comprehensive experiments showing OSoRA achieves comparable or superior performance to state-of-the-art methods with fewer parameters
\end{itemize}

Our work makes LLMs adaptation more accessible and efficient, 
enabling fine-tuning of large models on limited computational resources without sacrificing adaptation quality.

\section{Related Work}
PEFT began with inserting small adapter modules into each transformer block \citep{houlsby2019parameter}.
Concurrently, prompt-based methods such as Prompt-Tuning \citep{lester-etal-2021-power}, 
P-Tuning \citep{liu-etal-2022-p}, and P-Tuning v2 \citep{liu2021p} showed 
that a handful of continuous tokens prepended to the input can steer frozen language models toward new tasks while keeping all backbone weights intact.
These two lines established the principle that high-capacity language models can often be adapted with orders-of-magnitude fewer trainable parameters than full fine-tuning.

LoRA \citep{lora2022} popularized the idea of constraining weight updates to a rank-$r$ product of two small matrices, 
reducing trainable parameters from $O(dk)$ to $O(r(d+k))$ and sparing most optimizer state.
On top of this foundation, AdaLoRA \citep{adaptive2023} allocates rank budget across layers on the fly, 
and QLoRA \citep{qlora2023} combines LoRA with 4-bit quantization so that both training and inference fit on consumer GPUs.
VeRA \citep{vera2024} keeps the low-rank bases frozen and learns only two scaling vectors, 
achieving the same $r+d$ trainable parameters as our method while introducing variance-preserving random projections that improve generalization.

Several works seek more informative update directions than random bases.
DoRA \citep{dora2024} fine-tunes the norm of each weight and updates its direction, improving stability.
PiSSA \citep{pissa2024} leverages SVD to decompose weight matrices and selectively updates only the principal singular values and their corresponding vectors, 
preserving the model's inherent knowledge while enabling effective adaptation.

OSoRA unifies the advantages of the two branches above.
Like LoRA and VeRA, it constrains updates to a low-rank form and requires only $r+d$ trainable scalars, preserving memory and computational efficiency.
Unlike methods that rely on random or learned bases, OSoRA initializes its subspace with the top-$r$ singular vectors of the pretrained weights, 
capturing the model's dominant variation directions from the outset.
It further introduces two learnable vectors that can be transformed into diagonal matrices - one over output dimensions 
and one over rank components.
This synthesis yields a PEFT method that maintains LoRA's simplicity, matches VeRA's parameter efficiency, 
and inherits the informed initialization benefits demonstrated by PiSSA.

\section{Method}

\begin{figure*}
  \centering
  \adjustbox{max width=\linewidth, clip}{
    \includegraphics{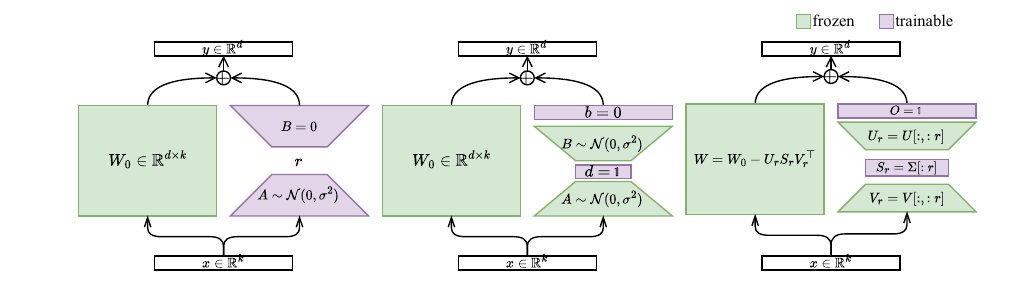}
  }
  \caption{Schematic comparison of LoRA (left), VeRA (middle) and OSoRA (right).
  LoRA adapts pretrained weights $W_0\in\mathbb{R}^{d\times k}$ by training low-rank matrices $A\in\mathbb{R}^{r\times k}$ and $B\in\mathbb{R}^{d\times r}$. 
  VeRA keeps these matrices frozen but introduces learnable scaling vectors $d\in\mathbb{R}^{r}$ and $b\in\mathbb{R}^{d}$.
  OSoRA applies SVD to decompose $W_0$ into singular vectors $U_r\in\mathbb{R}^{d\times r}$ and $V_r\in\mathbb{R}^{k\times r}$ 
  with corresponding singular values $S_r\in\mathbb{R}^{r}$. During fine-tuning, only $S_r$ and a learnable all-ones vector $O\in\mathbb{R}^{d}$ 
  are updated, while the singular vector matrices remain fixed.
  }
  \label{fig:compare}
\end{figure*}

In this section, 
we introduce Output-Dimension and Singular-Value Initialized Low-Rank Adaptation (OSoRA), 
a novel approach for efficient fine-tuning of pre-trained models. 
OSoRA builds upon and extends state-of-the-art methods such as VeRA \citep{vera2024} and LoRA \citep{lora2022}.
The key innovation of OSoRA is the strategic reparameterization of low-rank matrices using SVD.
Specifically, we maintain frozen pairs of matrices derived from singular vectors, 
while only updating the singular value vectors and a single output-dimension vector initialized as all-ones during training, 
as illustrated in Figure \ref{fig:compare}.
Like VeRA and LoRA, OSoRA allows the trained vectors and low-rank matrices to be seamlessly merged into the original weights,
eliminating any additional computational overhead during inference.

\subsection{Preliminaries}
LoRA fine-tunes LLMs using a product of two low-rank matrices $B\in\mathbb{R}^{d\times r}$ and $A\in\mathbb{R}^{r\times k}$.
For a pretrained weight matrix $W_0\in\mathbb{R}^{d\times k}$, 
LoRA constrains the weight update $\Delta W$ to a low-rank decomposition, as shown in Eq. \eqref{eq:lora}:
\begin{equation}
  y = W_0x + \Delta Wx = W_0x + \underline{BA}x
  \label{eq:lora}
\end{equation}
where underlined parameters indicate trainable components.
This approach allows the original weight matrix $W_0$ to remain frozen 
while only optimizing the low-rank matrices $A$ and $B$.
Since $r \ll \min(d,k)$, these matrices contain significantly fewer parameters than the original weight matrix,
making the fine-tuning process computationally efficient.

Building upon LoRA, VeRA further reduces parameter count and can be formulated as:
\begin{equation}
  y = W_0x + \Delta Wx = W_0x + \underline{\Lambda_b}B\underline{\Lambda_d}Ax
  \label{eq:vera}
\end{equation}
where $\underline{\Lambda_b}$ and $\underline{\Lambda_d}$ are diagonal matrices constructed from learnable vectors $b \in \mathbb{R}^d$ and $d \in \mathbb{R}^r$, respectively. 
Unlike LoRA, VeRA uses frozen, randomly initialized matrices $B \in \mathbb{R}^{d \times r}$ and $A \in \mathbb{R}^{r \times k}$, 
with adaptation occurring solely through the scaling vectors.

\subsection{Method Formulation}
OSoRA performs SVD to decompose the pretrained weight matrix $W_0$, as shown in Eq. \eqref{eq:osora}:
\begin{equation}
  W_0 = U\Sigma V^\top
  \label{eq:osora}
\end{equation}
where $U \in \mathbb{R}^{d \times d}$ and $V \in \mathbb{R}^{k \times k}$ are orthogonal matrices containing the left and right singular vectors of $W_0$, respectively, 
and $\Sigma \in \mathbb{R}^{d \times k}$ is a diagonal matrix containing the singular values of $W_0$ in descending order.

OSoRA selectively adapts only the top $r$ singular values and introduces a learnable scaling vector, 
while keeping the corresponding singular vectors fixed. 
The adaptation can be formulated as:
\begin{equation}
  y = W_0'x + \Delta Wx = W_0'x + \underline{\Lambda_O}U_r\underline{\Lambda_{S_r}}V_r^\top x
  \label{eq:osora_adapt}
\end{equation}
where $\underline{\Lambda_O} \in \mathbb{R}^{d \times d}$ is a diagonal matrix constructed from a learnable scaling vector $O \in \mathbb{R}^{d}$ (initialized as all-ones), 
$U_r \in \mathbb{R}^{d \times r}$ and $V_r \in \mathbb{R}^{k \times r}$ are the fixed left and right singular vectors corresponding to the top $r$ singular values, 
and $\underline{\Lambda_{S_r}} \in \mathbb{R}^{r \times r}$ is a diagonal matrix constructed from the learnable singular values $S_r \in \mathbb{R}^{r}$.
$W_0'$ represents the frozen component of the weight matrix after excluding the contribution of the top $r$ singular values and the corresponding singular vectors, 
which can be written as:
\begin{equation}
  W_0' = W_0 - \underline{\Lambda_O}U_r\underline{\Lambda_{S_r}}V_r^\top
\end{equation}

\subsection{Memory and Computational Considerations}
While OSoRA significantly reduces the number of trainable parameters to just $r + d$ during fine-tuning, 
it's important to clarify the overall memory footprint during training. 
Although only the singular values $S_r \in \mathbb{R}^r$ and the scaling vector $O \in \mathbb{R}^d$ are learnable, 
the method still requires storing the frozen singular vectors $U_r \in \mathbb{R}^{d \times r}$ and $V_r \in \mathbb{R}^{k \times r}$ in memory during training. 
These matrices contain $dr + kr$ elements, which is comparable to the memory requirements of LoRA and VeRA.

The total memory footprint during training can be expressed as:
\begin{equation}
  \mathcal{M}_{\text{OSoRA}} = (r + d) + (dr + kr) 
\end{equation}
where the first term $(r + d)$ represents the trainable parameters, 
and the second term $(dr + kr)$ represents the frozen singular vectors that must be stored in memory.

This clarification is important because while the trainable parameter count is significantly reduced, 
the overall memory and computational requirements during training remain similar to other low-rank adaptation methods. 
However, the key advantage of OSoRA is that after training, the adapted weights can be computed and merged into a single matrix:
\begin{equation}
  W = W_0' + \underline{\Lambda_O} U_r \underline{\Lambda_{S_r}} V_r^\top
\end{equation}

This means that while the singular vectors $U_r$ and $V_r$ need to be kept in memory during training, 
they do not need to be saved when storing checkpoints or the final adapted weights, 
significantly reducing storage requirements. 
During inference, only the merged weight matrix $W$ is needed, 
eliminating any additional memory or computational overhead compared to using the original pretrained weights.

\subsection{Necessity of Dual Vectors $O$ and $S_r$}
% A natural question arises: why do we need both the scaling vector $O \in \mathbb{R}^d$ and the singular value vector $S_r \in \mathbb{R}^r$? 
These vectors serve distinct purposes and operate in different dimensions:
\paragraph{$O \in \mathbb{R}^d$} controls scaling along the output dimension, 
allowing the model to selectively emphasize or de-emphasize specific output features.
\paragraph{$S_r \in \mathbb{R}^r$} controls the importance of each rank component, 
effectively weighting the contribution of each singular vector pair.

Since $r \ll d$ in typical applications (e.g., $r=256$ while $d=4096$), 
these vectors operate in spaces of different dimensionality and cannot be collapsed into a single vector. 
This dual-vector approach provides OSoRA with greater expressivity.
% it can simultaneously control which output dimensions are most adaptable (via $O$) and which directions in the weight space are most important for adaptation (via $S_r$).

Furthermore, initializing $S_r$ with the top singular values from the pretrained weights provides OSoRA with a principled starting point 
that captures the most important directions of variation in the original weight matrix, 
while $O$ allows for fine-grained control over how these directions affect each output dimension. 
This effectively enables fine-tuning within the most important low-rank subspace, 
while $O$ is responsible for regulating energy distribution across the complete output space.

\subsection{Parameter Efficiency Analysis}
OSoRA achieves significant parameter efficiency compared to other methods. 
The total number of trainable parameters in OSoRA is $r + d$, 
where $r$ is the rank and $d$ is the output dimension of the weight matrix.

\paragraph{Comparison with LoRA} LoRA requires $r(d+k)$ trainable parameters, 
where $k$ is the input dimension. 
The ratio of parameters between OSoRA and LoRA is:
\begin{equation}
  \frac{\mathcal{P}_{\text{OSoRA}}}{\mathcal{P}_{\text{LoRA}}} = \frac{r + d}{r(d+k)} = \frac{1}{d+k} + \frac{d}{r(d+k)}
\end{equation}
For large values of $r$, $d$, and $k$ (typical in LLMs), this ratio becomes very small, 
demonstrating OSoRA's superior parameter efficiency.

\paragraph{Comparison with VeRA} VeRA requires $r + d$ trainable parameters, the same as OSoRA. 
However, OSoRA's initialization from the pretrained weights' SVD provides a more informed starting point for fine-tuning, 
potentially leading to better performance with the same parameter count.

\subsection{Optimization Dynamics}
The optimization dynamics of OSoRA differ from those of other methods due to its unique parameterization. 
When updating the singular values $S_r$ and the scaling vector $O$, 
the gradients flow through the fixed singular vectors $U_r$ and $V_r$, 
which capture the principal directions of variation in the original weight matrix.

Let $\mathcal{L}$ be the loss function. The gradients with respect to the trainable parameters are:
\begin{equation}
  \frac{\partial \mathcal{L}}{\partial S_r} = \text{diag}(U_r^\top \Lambda_O \frac{\partial \mathcal{L}}{\partial \Delta W} V_r)
\end{equation}
\begin{equation}
  \frac{\partial \mathcal{L}}{\partial O} = \text{diag}(\frac{\partial \mathcal{L}}{\partial \Delta W} U_r \Lambda_{S_r} V_r^\top)
\end{equation}

These gradients show that the updates to $S_r$ are influenced by 
how well the corresponding singular vectors align with the desired weight update direction, 
while updates to $O$ are influenced by the overall contribution of each output dimension to the loss.

\begin{table*}[t]
  \centering
  \setlength{\tabcolsep}{1.5pt}
  \begin{tabular}{ccccccccccc}
    \toprule
    \textbf{Model} & \textbf{Method} & \textbf{BoolQ} & \textbf{PIQA} & \textbf{SIQA} & \textbf{HellaSwag} & \textbf{WinoGrande} & \textbf{ARC\_e} & \textbf{ARC\_c} & \textbf{OBQA} & \textbf{Avg.} \\
    \midrule
    \multirow{4}{*}{LLaMA2-13B} & LoRA & 65.84 & 73.78 & 53.68 & 48.63 & 51.78 & 79.01 & 59.32 & 60.00 & 61.51 \\
    & DoRA & 61.07 & \textbf{73.94} & 54.25 & 49.98 & 51.46 & 79.19 & \textbf{61.02} & 60.40 & 61.41 \\
    & PiSSA & 66.09 & 70.18 & 45.39 & 51.94 & \textbf{52.33} & \textbf{82.19} & 59.32 & \textbf{62.40} & 61.23 \\
    & VeRA & 67.16 & 67.63 & 48.41 & 48.78 & 51.85 & 79.19 & 55.25 & 57.80 & 59.38  \\
    &\cellcolor{gray!20} OSoRA & \cellcolor{gray!20}\textbf{74.10} & \cellcolor{gray!20}65.07 & \cellcolor{gray!20}\textbf{54.76} & \cellcolor{gray!20}\textbf{55.13} & \cellcolor{gray!20}50.83 & \cellcolor{gray!20}73.72 & \cellcolor{gray!20}50.51 & \cellcolor{gray!20}56.00 & \cellcolor{gray!20}60.02 \\
    
    \multirow{4}{*}{Qwen1.5-7B} & LoRA & 83.43 & 72.47 & 44.68 & 71.78 & 61.96 & 87.83 & 77.29 & 75.20 & 71.83 \\
    & DoRA & 83.24 & 70.95 & 44.68 & \textbf{71.82} & 61.88 & 88.01 & 77.29 & 76.00 & 71.73 \\
    & PiSSA & 84.04 & 74.32 & \textbf{44.73} & 71.53 & 61.64 & 87.65 & 78.31 & 74.20 & 72.05 \\
    & VeRA & 83.79 & 78.24 & 38.74 & 69.00 & \textbf{62.27} & 88.01 & 74.92 & 76.00 & 71.50 \\
    &\cellcolor{gray!20} OSoRA & \cellcolor{gray!20}\textbf{84.31} & \cellcolor{gray!20}\textbf{78.84} & \cellcolor{gray!20}38.84 & \cellcolor{gray!20}69.73 & \cellcolor{gray!20}61.56 & \cellcolor{gray!20}\textbf{88.54} & \cellcolor{gray!20}\textbf{78.64} & \cellcolor{gray!20}\textbf{76.20} & \cellcolor{gray!20}72.08 \\
    
    \multirow{4}{*}{Qwen2.5-32B} & LoRA & 89.85 & \textbf{90.75} & 46.16 & 92.11 & 79.16 & \textbf{97.53} & 93.90 & 89.40 & 84.86 \\
    & DoRA & \textbf{90.03} & 90.59 & 46.26 & 92.06 & 79.08 & 97.35 & \textbf{93.56} & 89.80 & 84.84 \\
    & PiSSA & 89.76 & 89.61 & \textbf{46.62} & 91.91 & 77.66 & 97.53 & 91.86 & 88.60 & 84.19 \\
    & VeRA & 87.37 & 84.87 & 43.04 & 92.67 & \textbf{80.66} & 95.41 & 90.85 & 89.20 & 83.00 \\
    &\cellcolor{gray!20} OSoRA & \cellcolor{gray!20}88.10 & \cellcolor{gray!20}85.85 & \cellcolor{gray!20}43.04 & \cellcolor{gray!20}\textbf{92.76} & \cellcolor{gray!20}78.69 & \cellcolor{gray!20}96.83 & \cellcolor{gray!20}89.15 & \cellcolor{gray!20}\textbf{91.40} & \cellcolor{gray!20}83.23 \\
    \bottomrule
  \end{tabular}
  \caption{Accuracy comparison of LLaMA2-13B, Qwen1.5-7B, and Qwen2.5-32B with different PEFT methods on eight commonsense reasoning tasks. The best results are highlighted in bold.}
  \label{tab:exp_commonsense_results}
\end{table*}

\section{Experiments}
\label{sec:experiments}
In this section, we present a comprehensive evaluation of OSoRA through a series of experiments.
We first compare OSoRA against state-of-the-art PEFT methods including LoRA, VeRA, DoRA, 
and other baselines on Common Sense Reasoning and Mathematics benchmarks.
We then examine OSoRA's robustness across different rank configurations to assess its stability and performance characteristics.

Additionally, we perform detailed ablation studies to analyze the contribution of each component in our method,
with particular focus on how different initialization strategies affect the overall performance.

\subsection{Common Sense Reasoning}
We evaluate OSoRA on a comprehensive suite of benchmarks: 
BoolQ \citep{boolq2019}, PIQA \citep{piqa2019}, SIQA \citep{sap-etal-2019-social},
HellaSwag \citep{hellaswag2019}, Winogrande \citep{winogrande2021}, ARC\_e and ARC\_c \citep{think2018},
and OpenBookQA \citep{can2018}. 
We utilize three language models:
LLaMA2-13B Chat \citep{llama2023}, Qwen1.5-7B Chat \citep{qwen1.5}, and Qwen2.5-32B Instruct \citep{qwen252025},
and configure rank settings of 512, 256, and 1024 for these models, respectively.
The CommonSenseQA \citep{commonsenseqa2019} dataset is used for training all models,
and OpenCompass \citep{opencompass2023} is employed as the evaluation framework.
Following the approach of \citet{lora2022}, OSoRA is applied to the query and value projection matrices 
in each self-attention module. 
The optimal learning rates, training epochs, and other hyperparameters 
were determined through systematic tuning, with detailed configurations available in 
Table \ref{tab:exp_commonsense_settings}.

As demonstrated in Table \ref{tab:exp_commonsense_results}, 
OSoRA achieves competitive performance across all evaluated models. 
For Qwen1.5-7B, OSoRA achieves the highest average score (72.08\%) among all methods, 
outperforming LoRA (71.83\%), DoRA (71.73\%), PiSSA (72.05\%), and VeRA (71.50\%). 
OSoRA excels particularly on BoolQ (84.31\%), PIQA (78.84\%), ARC\_e (88.54\%), ARC\_c (78.64\%), and OBQA (76.20\%), 
achieving the best scores among all methods. 
For LLaMA2-13B, OSoRA shows strong performance on BoolQ (74.10\%), SIQA (54.76\%), and HellaSwag (55.13\%), 
while for Qwen2.5-32B, it performs best on HellaSwag (92.76\%) and OBQA (91.40\%). 
These results are notable given OSoRA's significantly reduced parameter count compared to other methods.

\subsection{Mathematics}\label{subsec:math}
For the mathematical task, 
we follow the experimental setup from \citet{pissa2024} and fine-tune the Mistral-7B Instruct v0.3 \citep{mistral2023}
and LLaMA3-8B Instruct \citep{llama2024} models. 
The training set is the MetaMathQA dataset \citep{metamath2024}
and the evaluation framework is also the OpenCompass \citep{opencompass2023}.
The hyperparameters are detailed in Table \ref{tab:exp_math_settings}.

As shown in Table \ref{tab:exp_math_results}, 
OSoRA demonstrates superior performance on the mathematical task across both models. 
For Mistral-7B v0.3, OSoRA achieves the highest scores on both MATH \citep{measuring2021a} (12.10\%) and GSM8K \citep{cobbe2021training} (54.81\%), 
outperforming the next best method PiSSA by 0.14\% and 1.82\% respectively. 
Similarly, for LLaMA3-8B, OSoRA attains the best results with 27.36\% on MATH and 78.85\% on GSM8K, 
surpassing LoRA by 0.20\% and 5.39\% respectively. 
The average performance gain of OSoRA over other methods is particularly notable (33.46\% for Mistral-7B and 53.11\% for LLaMA3-8B), 
while requiring significantly fewer trainable parameters compared to alternative approaches.

\begin{table}[t]
  \centering
  \setlength{\tabcolsep}{1.5pt}
  \begin{tabular}{ccccc}
    \toprule
    \textbf{Model} & \textbf{Method} & \textbf{MATH} & \textbf{GSM8K} & \textbf{Avg.} \\
    \midrule
    \multirow{4}{*}{Mistral-7B v0.3} & LoRA & 11.68 & 51.40 & 31.54 \\
    & DoRA & 11.78 & 51.55 & 31.67 \\
    & PiSSA & 11.96 & 52.99 & 32.48 \\
    & VeRA & 10.70 & 49.20 & 29.95 \\
    &\cellcolor{gray!20} OSoRA & \cellcolor{gray!20}\textbf{12.10} & \cellcolor{gray!20}\textbf{54.81} & \cellcolor{gray!20}\textbf{33.46} \\
    
    \multirow{4}{*}{LLaMA3-8B} & LoRA & 27.16 & 73.46 & 50.31 \\
    & DoRA & 26.60 & 73.39 & 50.00 \\
    & PiSSA & 26.38 & 74.45 & 50.42 \\
    & VeRA & 24.24  & 75.59  & 49.92 \\
    &\cellcolor{gray!20} OSoRA & \cellcolor{gray!20}\textbf{27.36} & \cellcolor{gray!20}\textbf{78.85} & \cellcolor{gray!20}\textbf{53.11} \\
    \bottomrule
  \end{tabular}
  \caption{Accuracy comparison of Mistral-7B v0.3 and LLaMA3-8B with different PEFT methods on MATH and GSM8K benchmarks. 
  The table shows percentage scores for each method, with OSoRA achieving the highest performance on both benchmarks across both models. 
  Results are based on 4-shot evaluation, with the best scores in each category highlighted in bold.}
  \label{tab:exp_math_results}
\end{table}

\subsection{Robustness of Different rank settings}
This section explores the impact of various rank configurations on OSoRA, VeRA and LoRA
by adjusting $r$ within the set \{64, 128, 256, 512\} for OSoRA and VeRA, and \{2, 4, 8, 16, 32, 64\} for LoRA, respectively.
The performance of the fine-tuned models was assessed on GPQA \citep{gpqa2024} benchmark
and the accuracy of the Qwen2-7B model on the GPQA Diamond task is reported.
The learning rate is set to $2e^{-5}$ for LoRA, $0.005$ for VeRA and OSoRA.
Additionally, the batch size is set to $1$ for all methods, training for 1 epoch with a warmup rate of 0.03, cosine learning rate schedule.

\begin{table}[t]
  \centering
  \begin{tabular}{cccccc}
    \toprule
    \textbf{$r$} & \textbf{LoRA} & \textbf{VeRA} & \textbf{OSoRA} \\
    \midrule
    2 & 29.80 & - & - \\
    4 & 28.28 & - & - \\
    8 & 30.30 & - & - \\
    16 & 28.79 & - & - \\
    32 & 36.87 & - & - \\
    64 & 32.83 & 31.31 & 31.82 \\
    128 & - & 27.78 & 30.81 \\
    256 & - & 33.84 & 31.31 \\
    512 & - & 29.80 & 35.86 \\
    \bottomrule
  \end{tabular}
  \caption{Accuracy comparison of Qwen2-7B model with different PEFT methods (LoRA, VeRA, and OSoRA) across various rank settings on the GPQA Diamond task. The results show how different rank values affect model performance.}
  \label{tab:exp_robustness}
\end{table}

As shown in Table \ref{tab:exp_robustness} and Figure \ref{fig:gpqa}, 
we observe that OSoRA demonstrates more stable performance across different rank settings compared to LoRA and VeRA. 
While LoRA achieves its peak performance at $r=32$ (36.87\%), its accuracy fluctuates significantly across different ranks. 
VeRA shows similar inconsistency, with its best performance at $r=256$ (33.84\%). 
In contrast, OSoRA maintains relatively consistent performance across lower ranks and achieves its highest accuracy at $r=512$ (35.86\%). 
Figure \ref{fig:gpqa} further illustrates that as rank increases, LoRA's parameter count grows exponentially, 
whereas both VeRA and OSoRA maintain a more efficient linear growth in parameter count. 
This demonstrates that OSoRA offers a better balance between performance and parameter efficiency, particularly at higher rank settings.

\subsection{Ablation Study}
\paragraph{Impact of Training Individual Components ($S_r$ or $O$)}
The importance of jointly training both components $S_r$ and $O$ in Equation \eqref{eq:osora_adapt} is first examined.
In this analysis, two simplified variants are considered: one where only $S_r$ is trained while $O$ remains fixed as an all-ones vector,
and another where only $O$ is trained while $S_r$ remains fixed at the initial singular values derived from the decomposition of $W_0$. The experimental setup from Section \ref{subsec:math} is maintained.

\begin{figure}[t]
  \centering
  \adjustbox{max width=\linewidth, clip}{
    \includegraphics{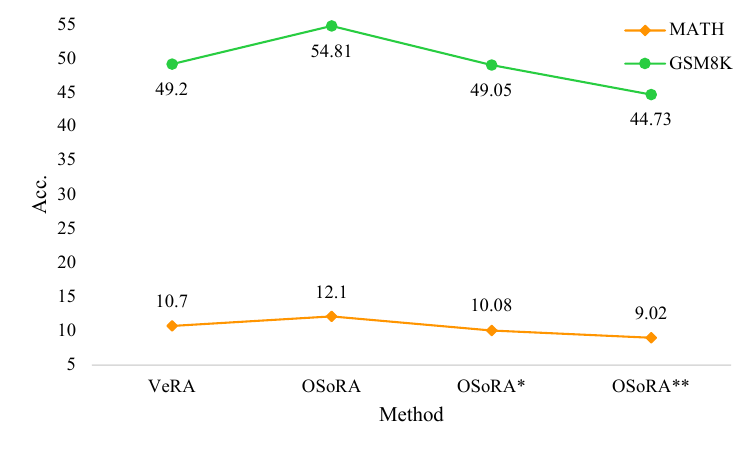}
  }
  \caption{Ablation study on the impact of training different components in OSoRA. 
  The figure compares accuracy on mathematical tasks (MATH and GSM8K) across three variants: 
  standard OSoRA with both $S_r$ and $O$ trained, OSoRA$^*$ with only $O$ trained (fixed $S_r$), 
  and OSoRA$^{**}$ with only $S_r$ trained (fixed $O$). 
  The results highlight that joint training of both components achieves the best performance, 
  while fixing the output dimension vector $O$ leads to the largest degradation in model accuracy.}
  \label{fig:ablation}
\end{figure}

The results of our ablation study on mathematical tasks (MATH and GSM8K) using the Mistral-7B v0.3 model are presented in Figure \ref{fig:ablation}. 
Three variants are compared: standard OSoRA (where both $S_r$ and $O$ are trained), OSoRA$^*$ (where $S_r$ is fixed and only $O$ is trained), 
and OSoRA$^{**}$ (where $O$ is fixed and only $S_r$ is trained).
It is clearly demonstrated by the results that superior performance is yielded by jointly training both components compared to when either component is trained individually. 
On the MATH benchmark, 12.1\% accuracy is achieved by standard OSoRA, by which OSoRA$^*$ (10.08\%) and OSoRA$^{**}$ (9.02\%) are significantly outperformed. 
Similarly, on GSM8K, 54.81\% accuracy is reached by standard OSoRA, compared to 49.05\% for OSoRA$^*$ and 44.73\% for OSoRA$^{**}$.

Notably, a more pronounced performance drop is observed when $O$ is fixed (OSoRA$^{**}$), 
by which it is suggested that a particularly crucial role in the adaptation process is played by the output dimension scaling vector. 
This finding is aligned with the theoretical understanding that fine-grained control over how the model's output dimensions are adjusted during adaptation is provided by $O$. 
Meanwhile, the importance of different principal components is modulated by the singular value vector $S_r$, by which optimal performance is also essentially enabled.

Our design choice to jointly train both components is validated by these results, 
as complementary aspects of the adaptation process are captured by them that cannot be fully realized when either component is trained in isolation.

\paragraph{Impact of Gaussian Distribution Initialization for Vector $O$}
In this experiment, the impact of initializing the learnable vector $O$ with $\mathcal{G}aussian$ distribution 
(denoted as OSoRA$_\mathcal{G}$) instead of ones in Equation \eqref{eq:osora_adapt} is investigated.

The results of this comparison on the mathematical tasks using LLaMA3-8B are presented in Table \ref{tab:exp_osora_gaussian}. 
It is revealed by the findings that notably worse performance is led to by initializing $O$ with a $\mathcal{G}aussian$ distribution (OSoRA$_\mathcal{G}$)
compared to the standard ones initialization used in OSoRA.
Specifically, only 24.12\% accuracy on MATH and 73.62\% on GSM8K are achieved by OSoRA$_\mathcal{G}$, compared to OSoRA's 27.36\% and 78.85\%, respectively.
A significant performance drop of 3.24\% on MATH and 5.23\% on GSM8K is represented by this.

Interestingly, comparable performance to VeRA is shown by OSoRA$_\mathcal{G}$ (24.12\% vs. 24.24\% on MATH and 73.62\% vs. 75.59\% on GSM8K),
by which it is suggested that a crucial role in OSoRA's effectiveness is played by the initialization strategy.
A more stable starting point for adaptation is provided by the all-ones initialization,
by which the pretrained weights' singular vectors can be leveraged more effectively from the beginning of training.

\begin{table}[t]
  \centering
  \begin{tabular}{cccc}
    \toprule
    \textbf{Model} & \textbf{MATH} & \textbf{GSM8K} & \textbf{Avg.}\\
    \midrule
    VeRA & 24.24 & 75.59 & 49.92 \\
    OSoRA & 27.36 & 78.85 & 53.11 \\
    OSoRA$_\mathcal{G}$ & 24.12 & 73.62 & 48.87 \\
    \bottomrule
  \end{tabular}
  \caption{Accuracy comparison of OSoRA and OSoRA$_\mathcal{G}$ on the MATH and GSM8K tasks. 
  The table shows that OSoRA achieves better performance than OSoRA$_\mathcal{G}$ on both tasks, with a 3.24\% higher accuracy on MATH (27.36\% vs. 24.12\%) and 5.23\% higher on GSM8K (78.85\% vs. 73.62\%), resulting in a 4.24\% higher average score (53.11\% vs. 48.87\%).}
  \label{tab:exp_osora_gaussian}
\end{table}

\paragraph{Exploring Input-Dimension Vector Adaptation: OSoRA$_k$}
In this experiment, OSoRA$_k$ is introduced as a variant of OSoRA where the learnable vector $O\in\mathbb{R}^d$ (output dimension)
in Equation \eqref{eq:osora_adapt} is replaced with $O\in\mathbb{R}^k$ (input dimension). The formulation can be expressed as:
\begin{equation}
  y=W_0'x+U_r\underline{\Lambda_{S_r}}V_r^\top\underline{\Lambda_O} x
  \label{eq:osora_adapt_k}
\end{equation}
where $\Lambda_O\in\mathbb{R}^{k\times k}$ is a diagonal matrix constructed from the learnable vector $O\in\mathbb{R}^k$.

Following the experimental setup described in Section \ref{subsec:math},
OSoRA$_k$ is evaluated against the original OSoRA on both MATH and GSM8K benchmarks.
The comparative results across different models are presented in Table \ref{tab:exp_osora_k}.
It is indicated by the findings that similar performance levels are achieved by both variants.
On Mistral-7B v0.3, a slight advantage on MATH (12.10\% vs. 11.98\%) is demonstrated by OSoRA,
while marginally better performance on GSM8K (55.88\% vs. 54.81\%) is shown by OSoRA$_k$.
The pattern is found to be consistent with LLaMA3-8B,
where a slight edge on MATH (27.36\% vs. 27.34\%) is maintained by OSoRA and a minimal advantage on GSM8K (78.92\% vs. 78.85\%) is shown by OSoRA$_k$.
Notably, approximately 50\% more trainable parameters (294,912 vs. 196,608) are required by OSoRA$_k$,
by which it is suggested that superior parameter efficiency is provided by the original OSoRA formulation while competitive performance is maintained.

\begin{table}
  \centering
  \setlength{\tabcolsep}{6pt}
  \begin{tabular}{cccc}
    \toprule
    \textbf{Method} & \textbf{Params} & \textbf{MATH} & \textbf{GSM8K}\\
    \midrule
    \rowcolor{cyan!10}\multicolumn{4}{c}{Mistral-7B v0.3} \\
    OSoRA & 196,608 & \textbf{12.10} & 54.81 \\
    OSoRA$_k$ & 294,912 & 11.98 & \textbf{55.88} \\
    \midrule
    \rowcolor{cyan!10}\multicolumn{4}{c}{LLaMA3-8B} \\
    OSoRA & 196,608 & \textbf{27.36} & 78.85 \\
    OSoRA$_k$ & 294,912 & 27.34 & \textbf{78.92}\\
    \bottomrule
  \end{tabular}
  \caption{Accuracy comparison of OSoRA and OSoRA$_k$ on the MATH and GSM8K tasks. 
  The table shows that OSoRA$_k$ achieves the comparable performance as OSoRA but with more parameters.}
  \label{tab:exp_osora_k}
\end{table}

\paragraph{Integrate OSoRA with DoRA}
The integration of OSoRA with DoRA is explored to investigate potential performance improvements from combining these PEFT methods. 
Weight updates are decomposed into magnitude and direction components by DoRA, 
while singular values with frozen singular vectors are optimized by OSoRA. 
The complementary strengths of both methods are leveraged through this combination.

The integration of OSoRA with DoRA can be formulated as:
\begin{equation}
  y=\underline{\|W_0\|_c}\frac{W_0'x+\underline{\Lambda_O}U_r\underline{\Lambda_{S_r}}V_r^\top x}{\|W_0'+\underline{\Lambda_O}U_r\underline{\Lambda_{S_r}}V_r^\top\|_c}
  \label{eq:osora_dora}
\end{equation}
where $\|\cdot\|_c$ denotes the vector-wise norm of a matrix across each column vector, similar to DoRA's approach.
OSoRA's parameter efficiency is maintained while DoRA's magnitude-direction decomposition benefits are gained through this formulation.

The combined approach is evaluated on the Mathematical task using the experimental setup described in Section \ref{subsec:math}.
Comparative results across different models are presented in Table \ref{tab:exp_osora_dora}.
Performance enhancement is indicated by integrating OSoRA with DoRA.
The best results on both MATH (12.36\%) and GSM8K (55.50\%) are achieved by the combined approach for Mistral-7B v0.3, outperforming both individual methods.
For LLaMA3-8B, while better MATH performance is shown by OSoRA alone, the highest GSM8K score (79.08\%) is achieved by the combined approach.
Only 360,448 trainable parameters are required by the combined approach,
which is significantly fewer than DoRA's 6,979,584 parameters,
by which OSoRA's parameter efficiency advantage is maintained while performance is potentially improved.

\begin{table}
  \centering
  \setlength{\tabcolsep}{3.5pt}
  \begin{tabular}{cccc}
    \toprule
    \textbf{Method} & \textbf{Params} & \textbf{MATH} & \textbf{GSM8K}\\
    \midrule
    \rowcolor{cyan!10}\multicolumn{4}{c}{Mistral-7B v0.3} \\
    DoRA & 6,979,584 & 11.78 & 51.55 \\
    OSoRA & 196,608 & 12.10 & 54.81 \\
    OSoRA + DoRA & 360,448 & \textbf{12.36} & \textbf{55.50} \\
    \midrule
    \rowcolor{cyan!10}\multicolumn{4}{c}{LLaMA3-8B} \\
    DoRA & 6,979,584 & 26.60 & 73.39 \\
    OSoRA & 196,608 & \textbf{27.36} & 78.85 \\
    OSoRA + DoRA & 360,448 & 27.12 & \textbf{79.08} \\
    \bottomrule
  \end{tabular}
  \caption{Accuracy comparison of DoRA, OSoRA, and their combination (OSoRA + DoRA) on the MATH and GSM8K tasks. 
  The table shows that combining OSoRA with DoRA can further improve performance while maintaining parameter efficiency.}
  \label{tab:exp_osora_dora}
\end{table}

\section{Conclusion}
In this paper, we introduced OSoRA, 
a novel PEFT method that performs SVD to adapt LLMs with minimal trainable parameters. 
Our approach combines the strengths of existing PEFT methods while addressing their limitations. 
By initializing with the top singular vectors of pretrained weights and training only singular values and scaling vectors, 
OSoRA achieves superior performance across various tasks while maintaining parameter efficiency.

Our extensive experiments demonstrate that OSoRA consistently outperforms state-of-the-art PEFT methods including LoRA, DoRA, PiSSA, 
and VeRA across common sense reasoning and mathematical tasks. 
The method's effectiveness is particularly notable on complex tasks like MATH and GSM8K, 
where it achieves comparable or better results with orders of magnitude fewer parameters than competing approaches.

We also explored variations of OSoRA, 
including OSoRA$_k$ with additional trainable parameters and integration with DoRA, 
showing the flexibility and extensibility of our approach. 
These results highlight the potential of informed initialization strategies in PEFT 
and contribute to making LLM fine-tuning more accessible and efficient, 
potentially enabling fine-tuning of increasingly large models on limited computational resources without sacrificing performance.

\section{Limitations}
Despite OSoRA's promising results, it faces several key limitations. The method requires computing SVD of pretrained weight matrices, introducing computational overhead that may challenge its use with extremely large models. Additionally, by operating within a fixed subspace defined by top singular vectors, OSoRA may struggle with tasks requiring significant departures from pretrained capabilities.

The performance heavily relies on appropriate rank selection - too small fails to capture important variations, while too large wastes computation. Unlike VeRA which can use any rank, OSoRA is constrained by the weight matrix dimensions. Our experiments also focused mainly on decoder-only models, leaving its effectiveness on other architectures like encoder-decoder or multimodal systems largely unexplored.

There are also concerns about potential overfitting on smaller datasets due to the concentrated adaptation in singular values and scaling vectors. Finally, integrating OSoRA with other PEFT methods introduces complexity in implementation and tuning that requires further investigation. These limitations point to important directions for future research and improvement.

% Entries for the entire Anthology, followed by custom entries
\bibliography{custom,ref}

\begin{thebibliography}{30}
\expandafter\ifx\csname natexlab\endcsname\relax\def\natexlab#1{#1}\fi

\bibitem[{Abacha et~al.(2025)Abacha, Yim, Fu, Sun, Yetisgen, Xia, and Lin}]{medec2025}
Asma~Ben Abacha, Wen-wai Yim, Yujuan Fu, Zhaoyi Sun, Meliha Yetisgen, Fei Xia, and Thomas Lin. 2025.
\newblock \href {https://doi.org/10.48550/arXiv.2412.19260} {{{MEDEC}}: {{A Benchmark}} for {{Medical Error Detection}} and {{Correction}} in {{Clinical Notes}}}.
\newblock \emph{arXiv preprint arXiv:2412.19260}.

\bibitem[{Bisk et~al.(2019)Bisk, Zellers, Bras, Gao, and Choi}]{piqa2019}
Yonatan Bisk, Rowan Zellers, Ronan~Le Bras, Jianfeng Gao, and Yejin Choi. 2019.
\newblock \href {https://doi.org/10.48550/arXiv.1911.11641} {{{PIQA}}: {{Reasoning}} about {{Physical Commonsense}} in {{Natural Language}}}.
\newblock \emph{arXiv preprint arXiv:1911.11641}.

\bibitem[{Brown et~al.(2020)Brown, Mann, Ryder, Subbiah, Kaplan, Dhariwal, Neelakantan, Shyam, Sastry, and Askell}]{language2020}
Tom Brown, Benjamin Mann, Nick Ryder, Melanie Subbiah, Jared~D Kaplan, Prafulla Dhariwal, Arvind Neelakantan, Pranav Shyam, Girish Sastry, and Amanda Askell. 2020.
\newblock \href {https://papers.nips.cc/paper_files/paper/2020/hash/1457c0d6bfcb4967418bfb8ac142f64a-Abstract.html} {Language {{Models}} are {{Few-Shot Learners}}}.
\newblock In \emph{Advances in {{Neural Information Processing Systems}}}, volume~33, pages 1877--1901. Curran Associates, Inc.

\bibitem[{Clark et~al.(2019)Clark, Lee, Chang, Kwiatkowski, Collins, and Toutanova}]{boolq2019}
Christopher Clark, Kenton Lee, Ming-Wei Chang, Tom Kwiatkowski, Michael Collins, and Kristina Toutanova. 2019.
\newblock \href {https://doi.org/10.18653/v1/N19-1300} {{{BoolQ}}: {{Exploring}} the {{Surprising Difficulty}} of {{Natural Yes}}/{{No Questions}}}.
\newblock In \emph{Proceedings of the 2019 {{Conference}} of the {{North American Chapter}} of the {{Association}} for {{Computational Linguistics}}: {{Human Language Technologies}}, {{Volume}} 1 ({{Long}} and {{Short Papers}})}, pages 2924--2936, Minneapolis, Minnesota. Association for Computational Linguistics.

\bibitem[{Clark et~al.(2018)Clark, Cowhey, Etzioni, Khot, Sabharwal, Schoenick, and Tafjord}]{think2018}
Peter Clark, Isaac Cowhey, Oren Etzioni, Tushar Khot, Ashish Sabharwal, Carissa Schoenick, and Oyvind Tafjord. 2018.
\newblock \href {https://doi.org/10.48550/arXiv.1803.05457} {Think you have {{Solved Question Answering}}? {{Try ARC}}, the {{AI2 Reasoning Challenge}}}.
\newblock \emph{arXiv preprint arXiv:1803.05457}.

\bibitem[{Cobbe et~al.(2021)Cobbe, Kosaraju, Bavarian, Chen, Jun, Kaiser, Plappert, Tworek, Hilton, Nakano et~al.}]{cobbe2021training}
Karl Cobbe, Vineet Kosaraju, Mohammad Bavarian, Mark Chen, Heewoo Jun, Lukasz Kaiser, Matthias Plappert, Jerry Tworek, Jacob Hilton, Reiichiro Nakano, et~al. 2021.
\newblock \href {https://arxiv.org/abs/2110.14168} {Training verifiers to solve math word problems}.
\newblock \emph{arXiv preprint arXiv:2110.14168}.

\bibitem[{Contributors(2023)}]{opencompass2023}
OpenCompass Contributors. 2023.
\newblock Open{Compass}: {A Universal Evaluation Platform for Foundation Models}.
\newblock \url{https://github.com/open-compass/opencompass}.

\bibitem[{Dettmers et~al.(2023)Dettmers, Pagnoni, Holtzman, and Zettlemoyer}]{qlora2023}
Tim Dettmers, Artidoro Pagnoni, Ari Holtzman, and Luke Zettlemoyer. 2023.
\newblock \href {https://doi.org/10.48550/arXiv.2305.14314} {{{QLoRA}}: {{Efficient Finetuning}} of {{Quantized LLMs}}}.
\newblock \emph{arXiv preprint arXiv:2305.14314}.

\bibitem[{Grattafiori et~al.(2024)Grattafiori, Dubey, Jauhri, Pandey, Kadian, {Al-Dahle}, Letman, Mathur, Schelten, and Vaughan}]{llama2024}
Aaron Grattafiori, Abhimanyu Dubey, Abhinav Jauhri, Abhinav Pandey, Abhishek Kadian, Ahmad {Al-Dahle}, Aiesha Letman, Akhil Mathur, Alan Schelten, and Alex Vaughan. 2024.
\newblock \href {https://doi.org/10.48550/arXiv.2407.21783} {The {{Llama}} 3 {{Herd}} of {{Models}}}.
\newblock \emph{arXiv preprint arXiv:2407.21783}.

\bibitem[{Hendrycks et~al.(2021)Hendrycks, Burns, Kadavath, Arora, Basart, Tang, Song, and Steinhardt}]{measuring2021a}
Dan Hendrycks, Collin Burns, Saurav Kadavath, Akul Arora, Steven Basart, Eric Tang, Dawn Song, and Jacob Steinhardt. 2021.
\newblock \href {https://openreview.net/pdf?id=7Bywt2mQsCe} {Measuring {{Mathematical Problem Solving With}} the {{MATH Dataset}}}.
\newblock In \emph{Thirty-Fifth {{Conference}} on {{Neural Information Processing Systems Datasets}} and {{Benchmarks Track}} ({{Round}} 2)}.

\bibitem[{Houlsby et~al.(2019)Houlsby, Giurgiu, Jastrzebski, Morrone, De~Laroussilhe, Gesmundo, Attariyan, and Gelly}]{houlsby2019parameter}
Neil Houlsby, Andrei Giurgiu, Stanislaw Jastrzebski, Bruna Morrone, Quentin De~Laroussilhe, Andrea Gesmundo, Mona Attariyan, and Sylvain Gelly. 2019.
\newblock Parameter-efficient transfer learning for nlp.
\newblock In \emph{International Conference on Machine Learning}, pages 2790--2799. PMLR.

\bibitem[{Hu et~al.(2022)Hu, Shen, Wallis, {Allen-Zhu}, Li, Wang, Wang, and Chen}]{lora2022}
Edward Hu, Yelong Shen, Phillip Wallis, Zeyuan {Allen-Zhu}, Yuanzhi Li, Shean Wang, Lu~Wang, and Weizhu Chen. 2022.
\newblock \href {https://openreview.net/pdf?id=nZeVKeeFYf9} {{{LoRA}}: {{Low-rank Adaptation}} of {{Large Language Models}}}.
\newblock In \emph{International {{Conference}} on {{Learning Representations}}}.

\bibitem[{Jiang et~al.(2023)Jiang, Sablayrolles, Mensch, Bamford, Chaplot, de~las Casas, Bressand, Lengyel, Lample, and Saulnier}]{mistral2023}
Albert~Q. Jiang, Alexandre Sablayrolles, Arthur Mensch, Chris Bamford, Devendra~Singh Chaplot, Diego de~las Casas, Florian Bressand, Gianna Lengyel, Guillaume Lample, and Lucile Saulnier. 2023.
\newblock \href {https://doi.org/10.48550/arXiv.2310.06825} {Mistral {{7B}}}.
\newblock \emph{arXiv preprint arXiv:2310.06825}.

\bibitem[{Kopiczko et~al.(2024)Kopiczko, Blankevoort, and Asano}]{vera2024}
Dawid~J Kopiczko, Tijmen Blankevoort, and Yuki~M Asano. 2024.
\newblock \href {https://openreview.net/pdf?id=NjNfLdxr3A} {{{VeRA}}: {{Vector-Based Random Matrix Adaptation}}}.
\newblock In \emph{International {{Conference}} on {{Learning Representations}}}.

\bibitem[{Lester et~al.(2021)Lester, Al-Rfou, and Constant}]{lester-etal-2021-power}
Brian Lester, Rami Al-Rfou, and Noah Constant. 2021.
\newblock \href {https://doi.org/10.18653/v1/2021.emnlp-main.243} {The power of scale for parameter-efficient prompt tuning}.
\newblock In \emph{Proceedings of the 2021 Conference on Empirical Methods in Natural Language Processing}, pages 3045--3059, Online and Punta Cana, Dominican Republic. Association for Computational Linguistics.

\bibitem[{Liu et~al.(2024)Liu, Wang, Yin, Molchanov, Wang, Cheng, and Chen}]{dora2024}
Shih-Yang Liu, Chien-Yi Wang, Hongxu Yin, Pavlo Molchanov, Yu-Chiang~Frank Wang, Kwang-Ting Cheng, and Min-Hung Chen. 2024.
\newblock \href {https://arxiv.org/pdf/2402.09353} {{{DoRA}}: {{Weight-Decomposed Low-Rank Adaptation}}}.
\newblock In \emph{Forty-First {{International Conference}} on {{Machine Learning}}}, pages 32100--32121. PMLR.

\bibitem[{Liu et~al.(2022)Liu, Ji, Fu, Tam, Du, Yang, and Tang}]{liu-etal-2022-p}
Xiao Liu, Kaixuan Ji, Yicheng Fu, Weng Tam, Zhengxiao Du, Zhilin Yang, and Jie Tang. 2022.
\newblock \href {https://doi.org/10.18653/v1/2022.acl-short.8} {{P}-tuning: Prompt tuning can be comparable to fine-tuning across scales and tasks}.
\newblock In \emph{Proceedings of the 60th Annual Meeting of the Association for Computational Linguistics (Volume 2: Short Papers)}, pages 61--68, Dublin, Ireland. Association for Computational Linguistics.

\bibitem[{Liu et~al.(2021)Liu, Ji, Fu, Tam, Du, Yang, and Tang}]{liu2021p}
Xiao Liu, Kaixuan Ji, Yicheng Fu, Weng~Lam Tam, Zhengxiao Du, Zhilin Yang, and Jie Tang. 2021.
\newblock \href {https://arxiv.org/abs/2110.07602} {P-tuning v2: Prompt tuning can be comparable to fine-tuning universally across scales and tasks}.
\newblock \emph{arXiv preprint arXiv:2110.07602}.

\bibitem[{Meng et~al.(2024)Meng, Wang, and Zhang}]{pissa2024}
Fanxu Meng, Zhaohui Wang, and Muhan Zhang. 2024.
\newblock \href {https://arxiv.org/pdf/2404.02948} {{{PiSSA}}: {{Principal Singular Values}} and {{Singular Vectors Adaptation}} of {{Large Language Models}}}.
\newblock In \emph{Advances in {{Neural Information Processing Systems}}}, volume~37, pages 121038--121072. Curran Associates, Inc.

\bibitem[{Mihaylov et~al.(2018)Mihaylov, Clark, Khot, and Sabharwal}]{can2018}
Todor Mihaylov, Peter Clark, Tushar Khot, and Ashish Sabharwal. 2018.
\newblock \href {https://doi.org/10.18653/v1/D18-1260} {Can a {{Suit}} of {{Armor Conduct Electricity}}? {{A New Dataset}} for {{Open Book Question Answering}}}.
\newblock In \emph{Proceedings of the 2018 {{Conference}} on {{Empirical Methods}} in {{Natural Language Processing}}}, pages 2381--2391, Brussels, Belgium. Association for Computational Linguistics.

\bibitem[{Qwen et~al.(2025)Qwen, Yang, Yang, Zhang, Hui, Zheng, Yu, Li, Liu, and Huang}]{qwen252025}
Qwen, An~Yang, Baosong Yang, Beichen Zhang, Binyuan Hui, Bo~Zheng, Bowen Yu, Chengyuan Li, Dayiheng Liu, and Fei Huang. 2025.
\newblock \href {https://doi.org/10.48550/arXiv.2412.15115} {Qwen2.5 {{Technical Report}}}.
\newblock \emph{arXiv preprint arXiv:2412.15115}.

\bibitem[{Rein et~al.(2024)Rein, Hou, Stickland, Petty, Pang, Dirani, Michael, and Bowman}]{gpqa2024}
David Rein, Betty~Li Hou, Asa~Cooper Stickland, Jackson Petty, Richard~Yuanzhe Pang, Julien Dirani, Julian Michael, and Samuel~R Bowman. 2024.
\newblock \href {https://arxiv.org/pdf/2311.12022} {{{GPQA}}: {{A Graduate-Level Google-Proof Q}}\&{{A Benchmark}}}.
\newblock In \emph{First {{Conference}} on {{Language Modeling}}}.

\bibitem[{Sakaguchi et~al.(2021)Sakaguchi, Bras, Bhagavatula, and Choi}]{winogrande2021}
Keisuke Sakaguchi, Ronan~Le Bras, Chandra Bhagavatula, and Yejin Choi. 2021.
\newblock \href {https://doi.org/10.1145/3474381} {{{WinoGrande}}: An adversarial winograd schema challenge at scale}.
\newblock \emph{Communications of the ACM}, 64(9):99--106.

\bibitem[{Sap et~al.(2019)Sap, Rashkin, Chen, Le~Bras, and Choi}]{sap-etal-2019-social}
Maarten Sap, Hannah Rashkin, Derek Chen, Ronan Le~Bras, and Yejin Choi. 2019.
\newblock \href {https://doi.org/10.18653/v1/D19-1454} {Social {IQ}a: Commonsense reasoning about social interactions}.
\newblock In \emph{Proceedings of the 2019 Conference on Empirical Methods in Natural Language Processing and the 9th International Joint Conference on Natural Language Processing (EMNLP-IJCNLP)}, pages 4463--4473, Hong Kong, China. Association for Computational Linguistics.

\bibitem[{Talmor et~al.(2019)Talmor, Herzig, Lourie, and Berant}]{commonsenseqa2019}
Alon Talmor, Jonathan Herzig, Nicholas Lourie, and Jonathan Berant. 2019.
\newblock \href {https://doi.org/10.18653/v1/N19-1421} {{{CommonsenseQA}}: {{A Question Answering Challenge Targeting Commonsense Knowledge}}}.
\newblock In \emph{Proceedings of the 2019 {{Conference}} of the {{North American Chapter}} of the {{Association}} for {{Computational Linguistics}}: {{Human Language Technologies}}, {{Volume}} 1 ({{Long}} and {{Short Papers}})}, pages 4149--4158, Minneapolis, Minnesota. Association for Computational Linguistics.

\bibitem[{Team(2024)}]{qwen1.5}
Qwen Team. 2024.
\newblock \href {https://qwenlm.github.io/blog/qwen1.5/} {Introducing {Qwen1.5}}.

\bibitem[{Touvron et~al.(2023)Touvron, Martin, Stone, Albert, Almahairi, Babaei, Bashlykov, Batra, Bhargava, and Bhosale}]{llama2023}
Hugo Touvron, Louis Martin, Kevin Stone, Peter Albert, Amjad Almahairi, Yasmine Babaei, Nikolay Bashlykov, Soumya Batra, Prajjwal Bhargava, and Shruti Bhosale. 2023.
\newblock \href {https://doi.org/10.48550/arXiv.2307.09288} {Llama 2: {{Open Foundation}} and {{Fine-Tuned Chat Models}}}.
\newblock \emph{arXiv preprint arXiv:2307.09288}.

\bibitem[{Yu et~al.(2024)Yu, Jiang, Shi, Yu, Liu, Zhang, Kwok, Li, Weller, and Liu}]{metamath2024}
Longhui Yu, Weisen Jiang, Han Shi, Jincheng Yu, Zhengying Liu, Yu~Zhang, James~T Kwok, Zhenguo Li, Adrian Weller, and Weiyang Liu. 2024.
\newblock \href {https://openreview.net/pdf?id=N8N0hgNDRt} {Metamath: {{Bootstrap Your Own Mathematical Questions}} for {{Large Language Models}}}.
\newblock In \emph{The {{Twelfth International Conference}} on {{Learning Representations}}}.

\bibitem[{Zellers et~al.(2019)Zellers, Holtzman, Bisk, Farhadi, and Choi}]{hellaswag2019}
Rowan Zellers, Ari Holtzman, Yonatan Bisk, Ali Farhadi, and Yejin Choi. 2019.
\newblock \href {https://doi.org/10.18653/v1/P19-1472} {{{HellaSwag}}: {{Can}} a {{Machine Really Finish Your Sentence}}?}
\newblock In \emph{Proceedings of the 57th {{Annual Meeting}} of the {{Association}} for {{Computational Linguistics}}}, pages 4791--4800, Florence, Italy. Association for Computational Linguistics.

\bibitem[{Zhang et~al.(2023)Zhang, Chen, Bukharin, He, Cheng, Chen, and Zhao}]{adaptive2023}
Qingru Zhang, Minshuo Chen, Alexander Bukharin, Pengcheng He, Yu~Cheng, Weizhu Chen, and Tuo Zhao. 2023.
\newblock \href {https://openreview.net/forum?id=lq62uWRJjiY} {Adaptive {{Budget Allocation}} for {{Parameter- Efficient Fine-Tuning}}}.
\newblock In \emph{International {{Conference}} on {{Learning Representations}}}.

\end{thebibliography}

\appendix

\section{Common Sense Reasoning Hyper-parameters}\label{sec:hyper-parameters}
This section details the hyper-parameter configurations used in our Common Sense Reasoning experiments, including learning rates, batch sizes, and rank settings across different model architectures.
\begin{table}[h]
  \centering
  \setlength{\tabcolsep}{4.5pt}
  \begin{tabular}{ccccc}
    \toprule
    \textbf{Model} & \textbf{Method} & \textbf{$r$} & \textbf{$\eta$} & \textbf{Batch$^\dagger$} \\
    \midrule
    \multirow{5}{*}{LLaMA2-13B} & LoRA & \multirow{3}{*}{16} & \multirow{3}{*}{2e-5} & \multirow{5}{*}{20} \\
    & DoRA & \\
    & PiSSA & \\
    & VeRA & \multirow{2}{*}{512} & \multirow{2}{*}{3e-3} & \\
    & OSoRA &  & \\
    
    \multirow{5}{*}{Qwen1.5-7B} & LoRA & \multirow{3}{*}{16} & \multirow{3}{*}{2e-5} & \multirow{3}{*}{20} \\
    & DoRA & \\
    & PiSSA & \\
    & VeRA & \multirow{2}{*}{256} & 5e-2 & 16 \\
    & OSoRA & & 3e-3 & 20 \\
    
    \multirow{5}{*}{Qwen2.5-32B} & LoRA & \multirow{3}{*}{16} & \multirow{3}{*}{2e-5} & \multirow{3}{*}{20} \\
    & DoRA & \\
    & PiSSA & \\
    & VeRA & \multirow{2}{*}{1024} & \multirow{2}{*}{8e-4} & \multirow{2}{*}{16} \\
    & OSoRA & \\
    
    \bottomrule
  \end{tabular}
  \caption{Hyper-parameters used for Common Sense Reasoning experiments. All methods were trained for 1 epoch with a warmup rate of 0.03, cosine learning rate schedule, and maximum sequence length of 512. $^\dagger$Batch represents the effective batch size (product of \textit{batch size} and \textit{gradient accumulation steps}).}
  \label{tab:exp_commonsense_settings}
\end{table}

\section{Mathematics Hyper-parameters}\label{sec:math_hyper-parameters}
This section details the hyper-parameter configurations used in our mathematical experiments and corresponding ablation studies.
\begin{table}[t]
  \centering
  \setlength{\tabcolsep}{4.5pt}
  \begin{tabular}{ccccc}
    \toprule
    \textbf{Model} & \textbf{Method} & \textbf{$r$} & \textbf{$\eta$} & \textbf{Batch$^\dagger$} \\
    \midrule
    \multirow{5}{*}{Mistral-7B v0.3} & LoRA & \multirow{3}{*}{16} & \multirow{3}{*}{2e-5} & \multirow{5}{*}{128} \\
    & DoRA & \\
    & PiSSA & \\
    & VeRA & \multirow{2}{*}{512} & 5e-3 \\
    & OSoRA & & 2e-5 \\
    \multirow{5}{*}{LLaMA3-8B} & LoRA & \multirow{3}{*}{16} & \multirow{3}{*}{2e-5} & \multirow{5}{*}{128} \\
    & DoRA & \\
    & PiSSA & \\
    & VeRA & \multirow{2}{*}{512} & 5e-3 \\
    & OSoRA & & 2e-5 \\
    \bottomrule
  \end{tabular}
  \caption{Hyper-parameters used for Mathematical experiments. All methods were trained for 1 epoch with a warmup rate of 0.03, cosine learning rate schedule, and maximum sequence length of 512. $^\dagger$Batch represents the effective batch size (product of \textit{batch size} and \textit{gradient accumulation steps}).}
  \label{tab:exp_math_settings}
\end{table}

\end{document}